\documentclass{article}

\usepackage[preprint]{neurips_2026}
\usepackage{tikz}
\usetikzlibrary{positioning, arrows.meta, calc} 

\usepackage[utf8]{inputenc} 
\usepackage[T1]{fontenc}    
\usepackage{hyperref}       
\usepackage{url}            
\usepackage{booktabs}       
\usepackage{tabularx}
\usepackage{multirow}
\usepackage{array} 

\usepackage{amsfonts}       
\usepackage{nicefrac}       
\usepackage{microtype}      
\usepackage{xcolor}         
\usepackage{graphicx}
\usepackage{enumitem}
\usepackage{amsmath}
\usepackage{subcaption}
\usepackage{xspace}
\usepackage{ragged2e}
\usepackage{letterspace}
\usepackage{multirow}
\usepackage{booktabs}
\usepackage{array}
\definecolor{darkblue}{HTML}{000090}
\usepackage{ragged2e} 
\usepackage[export]{adjustbox} 
\usepackage[most]{tcolorbox}
\AtBeginEnvironment{tcolorbox}{\small}

\hypersetup{
    colorlinks=true,
    citecolor=darkblue,
    linkcolor=darkblue,
    filecolor=darkblue,      
    urlcolor=darkblue,
}

\usepackage[textwidth=2cm]{todonotes}
\newif\ifdraft
\drafttrue
\ifdraft

    \newcommand{\marginvn}[1]{\todo[color=red!20,size=\tiny]{ VN:  #1}}
    \newcommand\inlinevn[1]{{\color{red} { \scriptsize (VN:#1)}}}

\else
    \newcommand\marginvn[1]{}
    \newcommand{\inlinevn}[1]{}

    \newcommand\note[1]{}
    \presetkeys{todonotes}{disable}{}
\fi

\title{The Topological Trouble With Transformers}
\author{%
  Michael C.~Mozer\\
  Google DeepMind\\
  \texttt{mcmozer@google.com} \\
  \And
  Shoaib Ahmed Siddiqui\\
  Google DeepMind \\
  \texttt{shoaibasidd@google.com} \\
  \And 
  Rosanne Liu\\
  Google DeepMind\\
  \texttt{rosanneliu@google.com}
}

\begin{document}

\maketitle
\newcommand{\faintred}[1]{\colorbox{red!20}{#1}}
\newcommand{\faintgreen}[1]{\colorbox{green!20}{#1}}

\begin{abstract}
Transformers encode structure in sequences via an expanding contextual history. However, their purely feedforward architecture fundamentally limits dynamic state tracking. State tracking---the iterative updating of latent variables reflecting an evolving environment---involves inherently sequential dependencies that feedforward networks struggle to maintain. Consequently, feedforward models push evolving state representations deeper into their layer stack with each new input step, rendering information inaccessible in shallow layers and ultimately exhausting the model's depth. While this depth limit can be bypassed by dynamic depth models and by explicit or latent thinking that externalizes state representations, these solutions are computationally and memory inefficient. In this article, we argue that temporally extended cognition requires refocusing from explicit thought traces to implicit activation dynamics via recurrent architectures. We introduce a taxonomy of recurrent and continuous-thought transformer architectures, categorizing them by their recurrence axis (depth versus step) and their ratio of input tokens to recurrence steps. Finally, we outline promising research directions, including enhanced state-space models and coarse-grained recurrence, to better integrate state tracking into modern foundation models.
\end{abstract}

\section{Introduction}


Progress in understanding human cognition has resulted from conceptualizing the
brain as a dynamical system.
In terms of its hardware, the physical brain is composed of billions of interacting neurons whose collective behavior is inherently dynamical. In terms of its function, the emergent mind can be usefully modeled as a dynamical process with a high-dimensional state, $s$, that evolves over time, modulated by external stimuli, $x$. These levels can be bridged by formalizing the state progression as $s_t= f(s_{t-1}, x_t)$, assuming discrete time $t$.

From this perspective, an ideal architecture for modeling temporally extended cognition would be a recurrent neural network (RNN), which explicitly performs such a state-update operation. In principle, gradient-based training procedures might discover the function $f$ from data such that the important input signals would be integrated into the state representation and held until later required.  The appeal of RNNs was somewhat dampened in the 1990s by the inherent limitations of gradient-based training \citep{mozer1992,hochreiter1998,hochreiter2001}.

Until the transformer \citep{Vaswanietal2017} came along, feedforward nets did not seem like a viable approach to replicating human thought and reasoning.
The transformer, with an audaciously long context window, retains all information in its history, often postponing the selection of relevant data until required for inference \citep{meng2022locating}.
In contrast, RNNs filter information as it arrives into a bottlenecked state
representation \citep{hochreiter1997long}. This article is about what can go wrong with the transformer's strategy and approaches that can address its limitations.

\section{State tracking}

Tracking the evolving world state is an essential ingredient for 
language understanding and 
reasoning, regardless of how tracking is achieved. The transformer's 
strategy often leverages its capacity to retrieve static, previously 
observed information from a context window. The attention mechanism of 
transformers is highly effective at retrieving past tokens \citep{olsson2022context}.
However,
this lookup mechanism is conceptually distinct from the explicit maintenance
of a dynamic state---the iterative, inherently sequential updating of 
latent variables that reflect a changing environment. The term 
\emph{belief state} is often used to refer to this compact, sufficient 
summary of the knowledge an AI agent has about its environment \citep{Chrisman1992,kaelbling1998planning}.
Belief state can be a set of facts or it can be a probability distribution over possible worlds.

To illustrate, consider the game of twenty questions. Each answer to a question
narrows the hypothesis space, and each subsequent question should be designed
to shrink the hypothesis space further. In a game where one is asked to
guess a number, if one is told that the number is larger than 50, it would 
make no sense to follow up it with a guess of 25. And on the flip side, 
if one is asked to think of a number between 1 and 100 and respond `higher'
or `lower' to guesses, maintaining the state---the hidden number---is critical 
to preventing inconsistent answers. Yet, here is a trace from Gemini 3 (Fast)
revealing a failure mode of models:\footnote{This example was originally suggested
by Gamal ElSayed circa 2020, still causing some modern foundation models to fail. Other models may have been trained to address this particular failure mode. 
\citet{baldelli2026} formally address the failure of standard models to reliably maintain a consistent hidden state over sequential interactions.}

\begin{tcolorbox}[sharp corners, colback=white, colframe=black, fontupper=\ttfamily]
User: Think of a number between 1 and 100 and I'll guess. Please respond only with "higher" or "lower" or "you got it". My first guess is 60.\newline
Model: lower\newline
User: 41\newline
Model: lower\newline
User: 70\newline
Model: \faintred{higher}
\end{tcolorbox}
Even though the model cannot generate a random number internally, it can 
play the game without actually having a number in mind simply by choosing 
responses consistent with its previous responses. However, consistency requires
tracking the valid range.  
Gemini 3 Thinking \emph{does} generate an explicit target, but then it fails to make use of the generated target even though it is also a part of the input token stream in this case:
\begin{tcolorbox}[sharp corners, colback=white, colframe=black, fontupper=\ttfamily]
User: Think of a number between 1 and 100 and I'll guess. Please respond only with "higher" or "lower" or "you got it". My first guess is 60.\newline
Model (thinking): \emph{I've begun designing the core game mechanics. I've successfully established the need for a number within the range of 1 to 100. The user's initial input, which is 60, is noted. Now I've internally selected the hidden number, choosing 42. Since 60 is higher than 42, my next step is to correctly communicate the response "lower".}\newline
Model: lower\newline
User: 42\newline
Model: \faintred{lower}
\end{tcolorbox}
Beyond games, state tracking is essential to understanding the ever-changing world,
the structure of arguments, and social interactions. State tracking failures 
in foundation models lead to loss of coherence in multi-turn conversations 
\citep{laban2025},  inefficiency in information gathering  \citep{sawyer2025}, 
and  breakdowns in communication and cooperation in multi-agent settings 
\citep{davidson2025,khatua2026}.

Without proper state tracking, models flip-flop in their
interpretations and fail to detect their inconsistencies, e.g.,
the meaning of a polysemous word \citep{lepori2025}:\footnote{The example
that follows was produced by Gemini 2.5 Flash in 2025. Although the model sometimes
responded correctly, and newer and more powerful models are much less susceptible to
this error, the example reveals a fundamental limitation of the core architecture.}
\begin{tcolorbox}[sharp corners, colback=white, colframe=black, fontupper=\ttfamily]
User: Fred took the day off work and pulled out his fishing pole. He drove to the bank. When he reaches the bank, should he wear boots or sandals?\newline
Model: While sandals may be more comfortable for a casual day, they are not the best choice for fishing, especially at a river bank. Here's why: ...\newline
User: Is it likely that Fred will find an ATM at this bank?\newline
Model: \faintred{It is highly likely that Fred will find an ATM at the bank.} Most banks, especially those located near bodies of water popular for fishing, have ATMs on-site for customer convenience.
\end{tcolorbox}

In this example, the model jumps from one interpretation of bank to the
other without a human-like acknowledgment of the reversal
(e.g., ``Oh wait, I misinterpreted. You must have been talking about a financial institution.''). 
We consider this flip-flop a failure to track the world situation, the model's earlier responses, and the 
listener's expectations.

While tracking Fred's location in the above example should not be challenging, in the most general case,
it is untenable for models---and people---to maintain and track probabilistic belief states over all environmental possibilities because the distributions explode in dimensionality.  People
adopt heuristics such as sampling \citep[e.g.,][]{Vul2014}, collapsing
distributions into prototypical cases \citep{Tversky1971}, or forming concrete
mental models most consistent with premises \citep{johnson1983}, 
like a MAP estimate. Nonetheless, even finite-memory, deterministic state
tracking can be unreliable in a transformer decoder.

\begin{figure}[b!]
    \centering
    \includegraphics[width=5.0in]{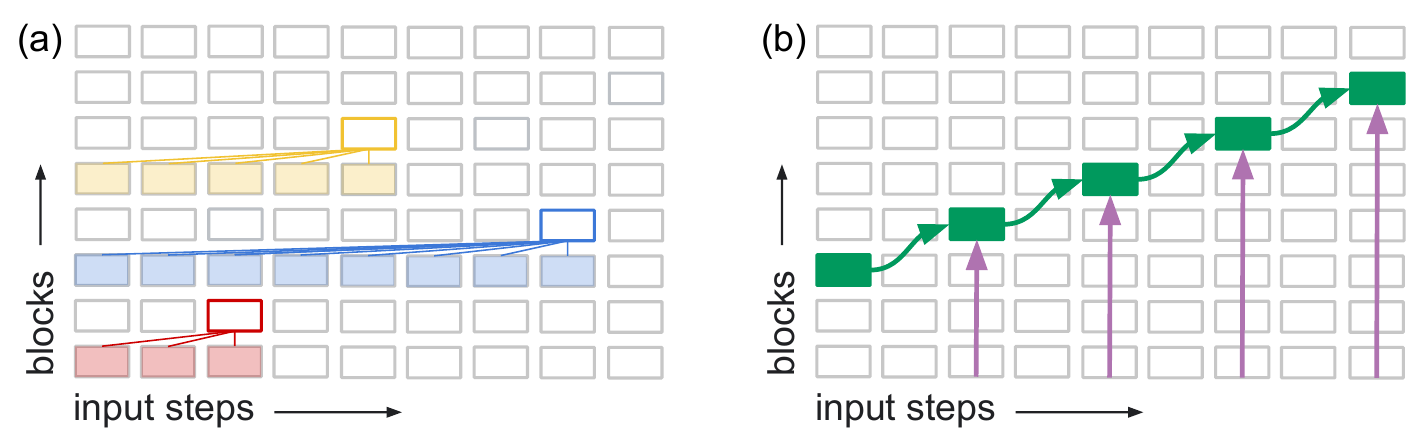}
    \caption{(a) Schematic depiction of transformer decoder architecture with input
    steps along horizontal axis and blocks (layers) along vertical axis. Activation
    flows upwards, from shallow to deep layers. For three selected blocks outlined in color, the 
    functional input connectivity of a causal model is shown by colored lines and shading of
    the input sources.
    (b) The propagation of state representations, indicated by the green rectangles.
    In cases where these representations depend on previous states and current inputs, 
    the state tracking capability of a feedforward transformer is fundamentally 
    limited by the depth of the model.
    } \label{fig:ff_transformer}
\end{figure}

\begin{figure}[tb]
    \centering
    \includegraphics[width=5.5in]{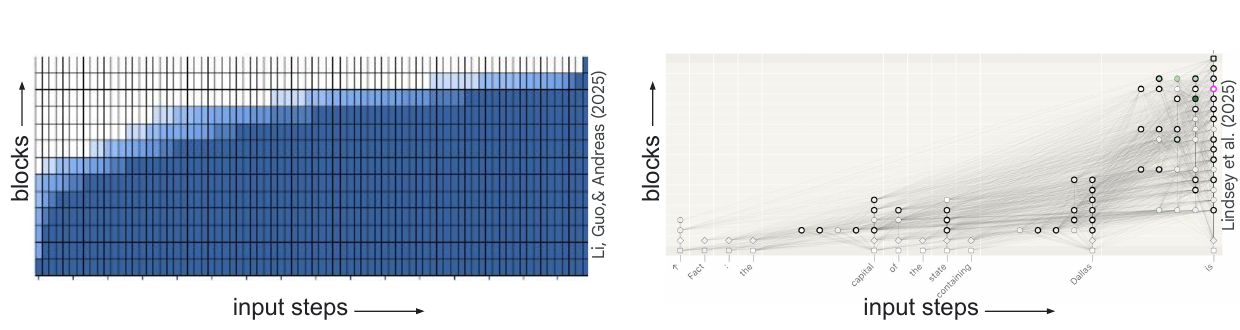}
    \caption{Two examples from the literature: 
    \citet{li2025} and \citet{Lindsey2025}.
    The details of each Figure are not critical, but in each case, the upward flow
    of information to deeper layers is depicted.
    } \label{fig:ff_other_models}
\end{figure}

We will use a schematic to explain the challenge of state tracking.
In Figure~\ref{fig:ff_transformer}a, we depict the transformer with input steps
shown along the horizontal axis and blocks (or layers) of the transformer along 
the vertical axis. Activation propagates from bottom to top (shallow to deep layers).
For three selected blocks, we use color to indicate the 
functional connectivity of a causal transformer: activation in a block is influenced
by all blocks immediately below and below to the left.
In Figure~\ref{fig:ff_transformer}b, we depict the flow of state information, indicated by the green rectangles. The integration
of the state representation and a new input (purple arrows) leads to a new
state representation, $s_t = f(s_{t-1},x_t)$. Because the architecture
is feedforward, $s_t$ must lie deeper in the stack than $s_{t-1}$, eventually
topping out of the model. Figure~\ref{fig:ff_other_models} shows examples from
several recent articles matching this upward activation flow. This flow
can make the integration of information over the sequence unreliable
\citep{biran2024,grant2025,lepori2025,sawyer2025,venhoff2025,baldelli2026}.

Not every state-tracking problem requires depth linear in the
number of layers. \citet{merrill2025} prove the necessity and sufficiency of 
$\log n$ layers to recognize regular language strings of up to length
$n$ and graph-connectivity problems with $n$ vertices. However, this proof
addresses only the constructability of solutions, not their learnability.
\citet{Fagnou2024} prove $\log n+1$ layers are required for an entity tracking
task with $n$ state changes and obtain empirical results consistent with this theory.

\begin{figure}[tb]
    \centering
    \includegraphics[width=4.25in]{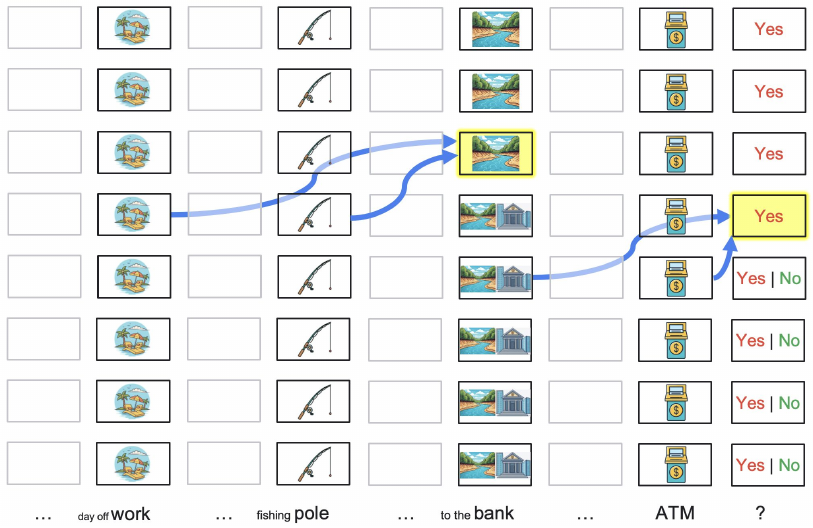}
    \caption{The depth of a state representation in a transformer can
    limit its utility for inference (adapted from \citeauthor{lepori2025}, \citeyear{lepori2025}).
    The input tokens are presented at the bottom, processed sequentially by the layers of the model.
    The symbols within the layers represent the model's internal belief state. This highlights 
    that despite the model converging to the correct belief (river bank) deep in the network 
    at the ``bank'' token, this high-level disambiguation is inaccessible to the shallower layers 
    of subsequent tokens. Consequently, when processing the final tokens, the model defaults 
    to superficial associations (money bank $\rightarrow$ ATM) early in the stack, resulting in an 
    incorrect prediction at the last token position.
    } \label{fig:bank}
\end{figure}

In practice, many researchers have identified clever solutions
obtained by training depth-limited models on specific finite sequence-length
problems \citep{li2025,piotrowski2025,prakash2026,shai2024}. Essentially, when
the state update function, $f$ is of a certain form, the sequence
of state updates may be composed into a simpler one-step function, e.g., 
there exists a function $g$ computable by a transformer layer such that
\[
s_t = f(\dots f(f(s_0, x_1), x_2), \dots, x_t) = g(s_0, x_1, \dots, x_t).
\]
Training losses have been proposed that aim to
steer models toward such solutions, to the extent they exist exactly or
approximately \citep{hu2025,teoh2025,huang2026semantictubepredictionbeating}.
However, when belief-state cascades push deeper and deeper into a network, computational limitations arise because the resulting representations are unavailable to shallower layers. To illustrate, Figure~\ref{fig:bank} depicts the processing of the bank dialog.
Using a technique called Patchscopes \citep{ghandeharioun2024}, \citet{lepori2025}
observed that the embedding of the polysemous word `bank' was ambiguous (i.e.,
a mixture of money bank vs. river bank) at shallow layers of the network, but 
deep in the network, the model selected the river bank interpretation. The
previous sentences provided context (e.g., `took the day off work', 'fishing pole') to support one interpretation over the other. In the Figure, we depict this contextualization as occurring at the sixth block of the transformer stack, 
which means that when subsequent tokens are processed, the disambiguation
is not available in blocks 1-5.  \citet{lepori2025} show that this delayed disambiguation leads to downstream errors whenever the response generation outpaces the model's internal semantic convergence, such as when the model forms its
yes/no response to the question about an ATM at the bank.

If reasonable models make errors when the inference cascade is
just two steps (\citeauthor{lepori2025} studied Gemma2-9B), it should be no surprise that models produce more severe failures
in comprehending extended multi-agent conversations. 

One solution to the depth dilemma is chain-of-thought style ``thinking'' where
the model recasts a deep representation as one or more output tokens, which are
then available to the model on its input \citep{wei2022chain}. For ordinary, step-by-step 
microcognition, this solution is a cop out. Inferences that people make 
automatically and unconsciously and then utilize consistently, such as the 
selection of a polysemous word's meaning, should not require elaborated, extended 
cognition. Even those who disagree with this desideratum should agree that if 
cognition in a transformer can be shifted from explicit thought traces to implicit
activation dynamics, the resulting model will be more powerful.

\section{Recurrent architectures}

As we previously mentioned, the state tracking ability of a feedforward model is limited
by model depth  \citep[e.g.,][]{merrill-sabharwal-2023-parallelism,Strobl2024,merrill2025}
and by the fact that effectively utilizing the state
representation becomes more challenging as it shifts upwards to deeper layers
\citep{biran2024,lepori2025,sawyer2025,venhoff2025}.

The alternative is a recurrent model, which is necessary to express arbitrary 
state dynamics, i.e., $s_t = f(s_{t-1},x_t)$. In this section, we explore
how to combine recurrence with transformers.
Given the subtle yet critical differences among varieties of recurrence, they are easily conflated and mistakenly
treated as equivalent.
This equivalence is problematic because recurrence is not
in itself sufficient for state tracking; the most popular
form of recurrence is unable to track state \citep{merrill2025illusion}.

\begin{figure}[t!]
    \centering
    \includegraphics[width=5in]{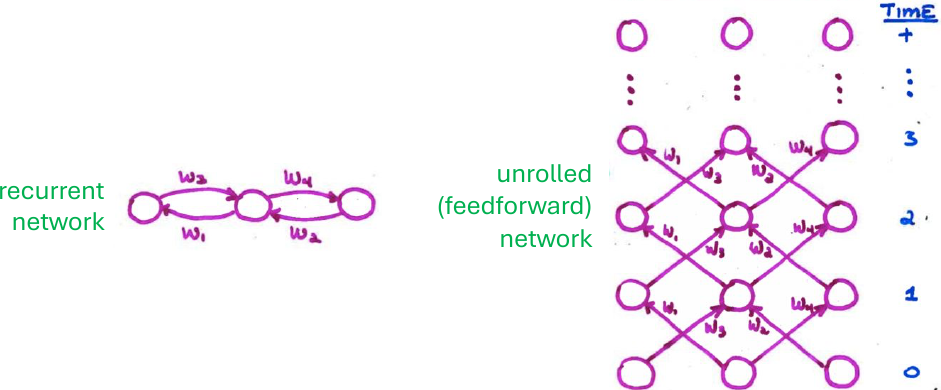}
    \caption{Unrolling a three-neuron recurrent neural net (left) into its equivalent feedforward network (right).
    Figure adapted from \citet{rumelhart1986}.
    } \label{fig:unrolling_generic}
\end{figure}

Figure~\ref{fig:unrolling_generic}, adapted from \citet{rumelhart1986}, shows a simple recurrent net (left) and 
the net  unrolled $t$ steps to form a weight-constrained feedforward net (right).
Unrolling the net requires making a copy of all neurons in the network for each
step of recurrence. At each step, activation flows through each connection.

\begin{figure}[b!]
    \centering
    \includegraphics[width=5.5in]{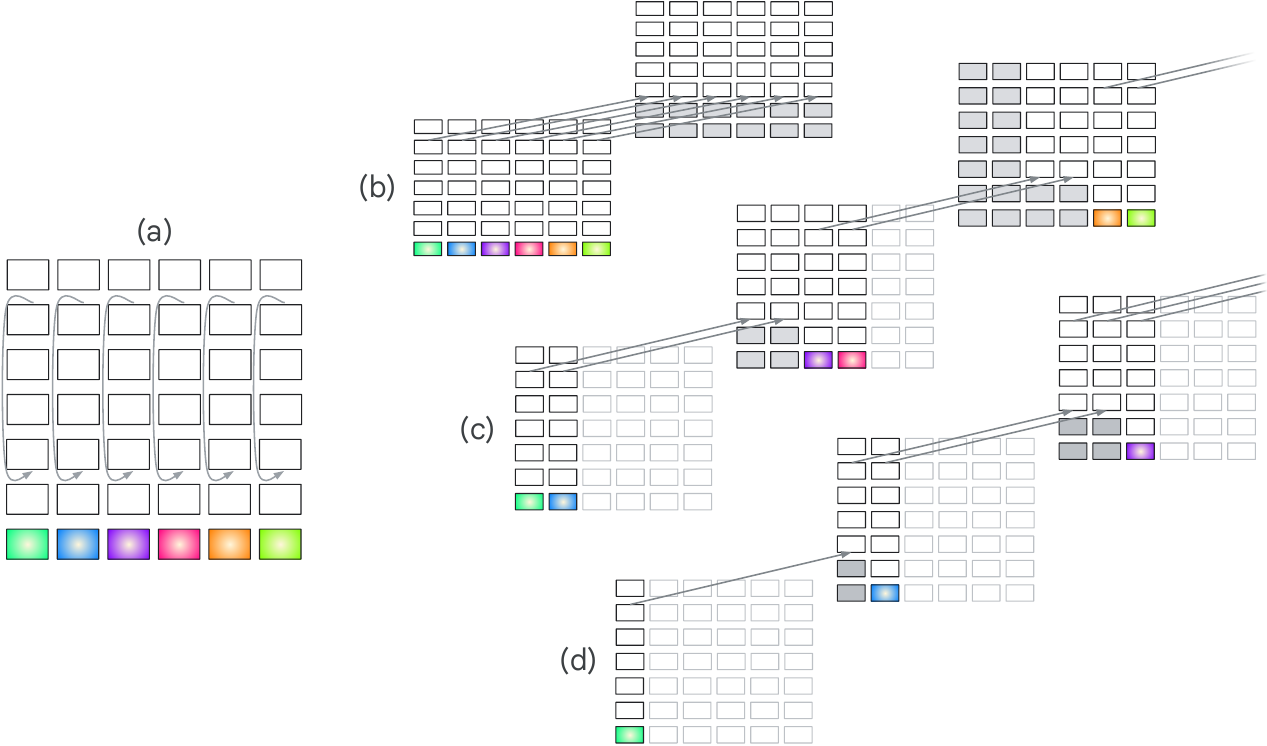}
    \caption{Unrolling a transformer. Each rectangle represents a transformer
    layer. The colored boxes represent distinct input tokens. The arrows indicate 
    flow of activity. Note that the figure represents the output of the full transformer layer, which includes both the self-attention as well as the feedforward layer.
    The gray filled rectangles indicate blocks that can
    be replaced by the frozen KV cache as their values no longer change. The 
    light grey unfilled rectangles represent input steps that are ignored at a given
    unrolling step. (a) A
    transformer with recurrence that allows activation from a deep layer to influence
    a shallow layer. (b) An unrolled depth-recurrent transformer.
    (c) An unrolled blockwise-recurrent transformer.
    (d) A transformer with potential attractor dynamics.
    } \label{fig:unrolling_transformer}
\end{figure}
Now consider a transformer with some recurrent connections, as depicted in
Figure~\ref{fig:unrolling_transformer}a. As before, each box denotes a 
transformer block and the horizontal and vertical axes correspond to input tokens 
and layers, respectively. The additional arrows are meant to illustrate
one possible type of recurrence involving activation flow from a deep layer to 
a shallow layer at every input step. The nature of the activation flow 
is not important for our purposes; for example, the shallow layer may generate 
queries that allow  it to cross attend to keys and values from the deep layer \citep[e.g.,][]{fan2021}.

Unlike the simple RNN in Figure~\ref{fig:unrolling_generic}, which can be unrolled 
in only one way, unrolling the recurrent transformer in 
Figure~\ref{fig:unrolling_transformer}a  results in ambiguity
due to the fact that that the transformer
architecture incorporates three distinct ordered dimensions:
(1) the layers of the architecture, bottom to top (shallow to deep) in the Figure;
(2) the input steps of the architecture, left to right in the Figure; and
(3) autoregressive steps performed at execution. During model pretraining,
an ordinary (feedforward) transformer has only a single autoregressive step 
because all input steps are run in parallel; and during inference, autoregressive
steps are typically confounded with input steps, although multi-token prediction
allows for multiple input/output steps per autoregressive step \citep{gloeckle2024better}.
However,
recurrence allows for further decoupling. In particular, many forms of
recurrence require that even when a model is trained via teacher forcing,
it must still be unrolled autoregressively \citep{teoh2025next}. This necessary sequentiality is
what we mean by \emph{autoregressive unrolling}, not
the sequentiality that arises from token-by-token generation in a pure feedforward model.

Figure~\ref{fig:unrolling_transformer}b depicts a transformer unrolled in
depth, sometimes referred to as a \emph{looped transformer}. 
Depth recurrence---whether
of individual layers or ranges of layers, and whether deterministic or adaptive---is 
a very popular and successful approach.
Some methods are designed and trained to allow for inference time scaling
\citep[e.g.,][]{yang2024,nowak2024,raposo2024,alabdulmohsin2025recursive,bae2025,chen2025,geiping2025,rodkin2025,yu2025backattn,zhu2025,zeng2026ponderlm,jeddi2026loopformer}; others incorporate recurrence via pretraining \citep{sanyal2026} or fine tuning a pretrained model \citep{koishekenov2025,mcleish2025}; and surprisingly, several operate purely as an
inference-time method to improve reasoning
\citep{li2025skiplayerloopit,chen2026,ng2026rys}.

While depth recurrence can increase the expressivity of a 
transformer \citep{saunshi2025}, it does not enable indefinite state tracking;
the propagation is still depth limited. To appreciate this fact, pick any layer $l$
in the lower or upper stack as the representation of $s(0)$, the state at the 
first stack. Then note that any $s(t+1)$, if it is to depend arbitrarily on 
$s(t)$ recursively, must be in a higher layer. The state representation still
shifts upward due to the parallel propagation of activation across steps $t$,
regardless of how deep the transformer is made to be with recurrent depth (Figure~\ref{fig:ff_transformer}b).

Indefinite tracking of state with an arbitrary state update function requires 
sequential dependency that precludes parallelization across the sequence length
\emph{during training}.  Two examples of autoregressive updates are presented in
Figures~\ref{fig:unrolling_transformer}c,d. Figure~\ref{fig:unrolling_transformer}c 
depicts a blockwise-recurrent model  \citep{hutchins2022,chevalier2023adaptinglanguagemodelscompress,chen2025melodi,borazjanizadeh2026} 
in which a subsequence of input steps  is run in parallel (two in the Figure) followed 
by an  autoregressive iteration; Figure~\ref{fig:unrolling_transformer}d 
shows a model in which one input step is presented per autoregressive step, and at
step $t$, all stacks up to $t-1$ send a signal from the deep layer to the shallow
layer, yielding a fully recurrent model.  This model may have attractor dynamics
since each layer continues to update, converging only when all previous steps have
reached asymptote.

\begin{figure}[b!]
    \centering
    \includegraphics[width=5.5in]{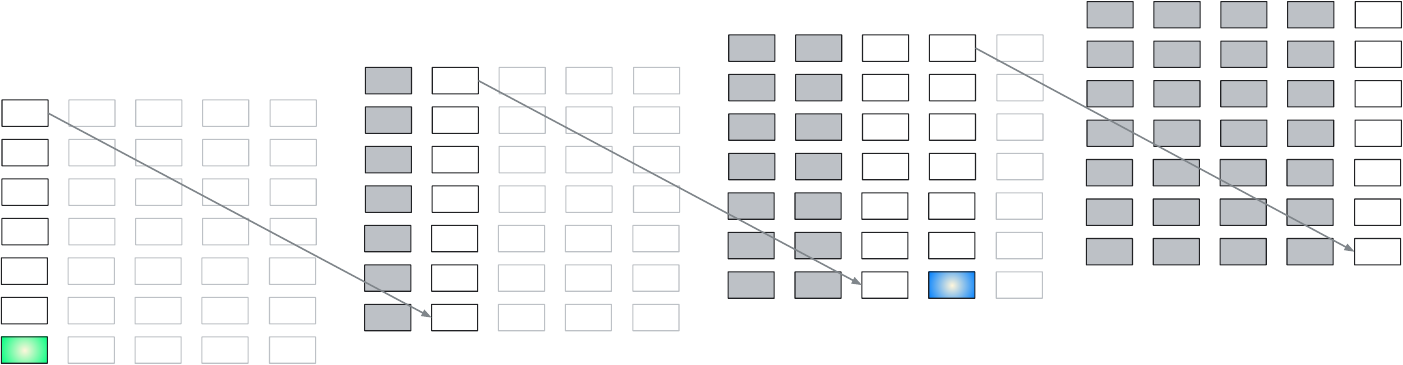}
    \caption{Unrolling a latent-thought model, where the model feeds back its latent thoughts as input to the model for multiple auto-regressive steps before processing the next input token, which is marked with the blue color \citep[e.g.,][]{hao2025coconut,jolicoeurmartineau2025}.
    } \label{fig:unrolling_latent_thinking}
\end{figure}
Having addressed recurrence with multiple input steps per autoregressive step (Figure~\ref{fig:unrolling_transformer}c) and a standard one-to-one mapping
(Figure~\ref{fig:unrolling_transformer}d), we can also consider a setup where
multiple autoregressive steps are executed for each input step
(Figure~\ref{fig:unrolling_latent_thinking}).
Latent-thought models have this form \citep[e.g.,][]{hao2025coconut,jolicoeurmartineau2025}.
Some of these models can track state, whereas others do not \citep[e.g.,][]{galashov2025}.

\begin{figure}[b!]
    \centering
    \includegraphics[width=5.5in]{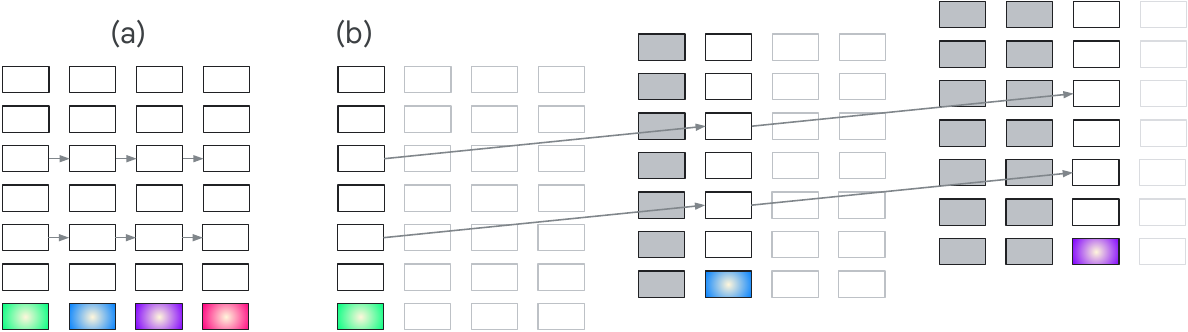}
    \caption{Unrolling an SSM, where information from the previous input step at layer $l$ flows directly to the next step in the horizontal direction \citep[e.g.,][]{allenzhu2025,gu2024}.
    } \label{fig:unrolling_ssm}
\end{figure}
The examples of recurrence we have given all involve signal propagation from
deeper layers to shallower layers, but within-layer propagation over inputs steps
(Figure~\ref{fig:unrolling_ssm}) also yields state dynamics \citep[e.g.,][]{allenzhu2025,gu2024,fagnou2026}.

\begin{table}[t!]
\centering
\caption{A taxonomy of recurrent transformer architectures, with selected examples. Note that ratio $> 1$ indicates processing multiple tokens per recurrence step, while ratio $< 1$ indicates multiple recurrence steps per token.}
\label{tab:recurrent_transformer_space}
\begin{tabularx}{5.5in}{|m{1.1in}|X|X|X|} 
\hline
\multirow{2}{*}{\textbf{Recurrence Axis}}
 & \multicolumn{3}{c|}{\textbf{Input Tokens Per Recurrence Step}} \\ \cline{2-4} 
& \multicolumn{1}{c|}{\textit{Ratio $>1$}} & \multicolumn{1}{c|}{\textit{Ratio $=1$}} & \multicolumn{1}{c|}{\textit{Ratio $<1$}} \\ \hline
\textit{Depth} &  
\tiny\RaggedRight
\color{brown}[Figure~\ref{fig:unrolling_transformer}b] \color{black}
looped transformer \citep{giannou2023looped},
universal transformer \citep{dehghani2018universal},
RINS \citep{alabdulmohsin2025recursive}
& 
\tiny\RaggedRight 
\color{brown}[Figure~\ref{fig:unrolling_transformer}d]\color{black}
~~Recirculation \citep{mozer2026recirculation}
& 
\tiny\RaggedRight
\\ \hline
\textit{Step} & 
\tiny\RaggedRight
block-recurrent transformers \citep{hutchins2022}
& 
\tiny\RaggedRight 
\color{brown}[Figure~\ref{fig:unrolling_ssm}b] \color{black}
linear attention \citep{Katharopoulos2020Transformers},
DeltaNet \citep{schlag2021},
MAMBA \citep{gu2024},
canon layers \citep{allenzhu2025}, 
PaTH attention \citep{yang2025path},
RWKV-7 \citep{peng2025rwkv},
test-time regression \citep{sun2025learning}
& 
\tiny\RaggedRight
DeltaProduct \citep{siems2025deltaproduct},
FSRM \citep{takashiro2026}
\\ \hline
\textit{Depth + Step} & 
\tiny\RaggedRight
\color{brown}[Figure~\ref{fig:unrolling_transformer}c] \color{black}
recurrent memory transformer \citep{bulatov2022},
RINs \citep{jabri2023},
sentence gestalt \citep{borazjanizadeh2026}
& 
\tiny\RaggedRight
feedback transformer \citep{fan2021}, full-bandwidth transformer  \citep{wang2026fullbandwidthtransformer}, T$^2$MLR \citep{cai2026t2mlrtransformertemporalmiddlelayer}
& 
\tiny\RaggedRight
\color{brown}[Figure~\ref{fig:unrolling_latent_thinking}] \color{black}
COCONUT \citep{hao2025coconut},
hierarchical reasoning model \citep{jolicoeurmartineau2025},
CYB \citep{galashov2025}
\\ \hline
\end{tabularx}
\end{table}

Table~\ref{tab:recurrent_transformer_space} attempts to lay out a taxonomy of
recurrent transformer architectures that includes the cases we have considered
so far. We characterize architectures along two dimensions: \emph{recurrence axis} and \emph{input tokens per recurrence step}.  The axis can be depth alone
(Figures~\ref{fig:unrolling_transformer}b,c,d), in step alone
(Figure~\ref{fig:unrolling_ssm}), and in both depth and step
(Figure~\ref{fig:unrolling_latent_thinking}).
The ratio of input tokens to recurrence steps can be greater than one
(Figure~\ref{fig:unrolling_transformer}b,c), equal to one 
(Figure~\ref{fig:unrolling_transformer}d, \ref{fig:unrolling_ssm}), or less
than one (Figure~\ref{fig:unrolling_latent_thinking}).\footnote{We define a `recurrence step' strictly as a sequential dependency that precludes parallelization across the sequence length during training.}
In cells of the 
taxonomy, we list some popular and representative transformer-based architectures. 
We omit architectures that are described as ``recurrent'' in the sense
that processing is iterative, but that are neither recurrent in depth nor
step, e.g., Transformer-XL \citep{dai2019transformerxlattentivelanguagemodels}.
Being recurrent in depth and/or step is necessary for state tracking but is not sufficient.
Essentially, full-fledged state tracking requires sequential dynamics during training; any model that
can be entirely parallelized across the context has limitations in updating state.

While we have not shown all such architectures, we have not succeeded in identifying
any examples of work that lies in the empty cells of the taxonomy. Some of these
cells would be worth exploration.
For example, architectures in the second row, third column, 
could include ones with within-layer attractor dynamics, where the model's 
activation may iterate to convergence before advancing
to the next token. Architectures in the first row, second and third columns, are
interesting because they feed activation from a deep layer to a shallow layer (once 
for $\text{ratio}=1$, multiple times for $\text{ratio}>1$) in a manner that---unlike
the looped  transformer---does allow for indefinite state propagation.

We have deliberately sidestepped the question of what a recurrent arrow signifies
in terms of activation dynamics. Different models make different proposals.
The arrow could indicate copying key-value cache \citep{yang2024kvsharer}, it could represent a source 
of keys and values for cross attention \citep{fan2021} or self attention \citep{oncescu2026}, or it might represent a direct connection through a linear layer,
an MLP, or an adapter such as LoRA \citep{hu2022lora}.
The coupling of step-to-step recurrence with attention---which gives neurons direct visibility to the 
entire  step history---prevents credit assignment bottlenecks that arise when training traditional
recurrent neural networks \citep{Ke2018,oncescu2026}.

\section{Architectural limitations and workarounds}

\citet{liu2026serialscalinghypothesis} point to the weakness of modern massively
parallel architectures on problems that are inherently sequential, problems where 
combinatorics make it impractical to parallelize, such as 
state tracking, multihop inference, and planning. Formal analyses point
to bounds on serial capacity for transformers 
\citep{merrill-sabharwal-2023-parallelism}. State-space models (SSMs) are often
touted as a means of state propagation, but SSMs with linear updates are
no more expressive than an ordinary transformer \citep{merrill2025illusion}.
In contrast, chain of thought does enhance model expressivity 
\citep{li2024chain,merrill2024cot}, as one would intuit:  Allowing a model
to talk to itself, whether in natural language or latent space, sends signals
from deep in the transformer to shallow layers, thereby propagating state forward.
It is no wonder that frontier models have become increasingly reliant on internal
`thinking'. However, the reliance on intermediate outputs to track
micro-state may perform wasteful computation steps and unnecessarily consume the context window. Implicit activation dynamics---albeit recurrent---might be adequate to 
efficiently update mundane state information of the sort that people can 
process unconsciously  and automatically. 
In human terms, it is fine to talk to yourself if you are reasoning through a 
calculus problem,  but it is a bit weird to have to continually remind yourself of 
the relationship between two characters in a book you are reading.

Given their limitations in state tracking, why are transformers as successful
as they are? The short answer is that by being able to reexamine their entire input
history, they can often turn a state-tracking problem into a working memory problem, 
i.e., into retrieval from the context window. For example, consider the
\emph{latch} problem from the 1990s that required a sequence-processing model latch
on to a bit of information early in a sequence and to retain it over a long time gap
\citep{mozer1991induction,Bengio1994}. This problem in large part motivated LSTM but is entirely trivial for a transformer that can re-index into its input history to retrieve
the early information. Transformers learn many clever strategies including this
sort of lookback \citep{prakash2026}, associatve scans \citep{li2025},
more specialized algorithms for formal language understanding
\citep{allenzhu2025physicslanguagemodels1,piotrowski2025}, including
belief updating that reflects state uncertainty \citep{shai2024}.
A simple illustration is the computation of parity from an input sequence
of $0$'s and $1$'s. Instead of maintaining a parity state, the model can compute pairwise
parity in the first layer of a transformer, then combine the pairs into four-bit
parity in the second layer, and so forth, achieving a computation from scratch in
$\log_2 n$ feedforward layers for a sequence of maximum length $n$.
Such alternate and even shortcut solutions are common for transformers to develop in practice \citep{liu2022transformers}.

Another factor in the success of transformers is their support of
\emph{state compositionality}.  State need not be---as we have depicted 
it in Figure~\ref{fig:ff_transformer}---a monolithic representation.
The representation of state can be split across embeddings and updated
asynchronously for each component. For example, if the model needs to 
track the changing locations of two entities, those state variables can be updated
independently for each entity.

\section{Promising directions}
In addition to the native mechanisms that transformers have to estimate state, 
several emerging research directions seem particularly promising for promoting
state maintenance and updating. These directions balance the need for 
expressivity with computational feasibility.

\subsection{Enhanced State-Space Models}
While most linear SSMs do not exceed the expressivity of standard transformers, 
Delta Net \citep{schlag2021}, when its eigenvalue range is extended to include negative 
values \citep{grazzi2025}, maintains the
highly desirable property of being trainable in parallel while simultaneously achieving
greater expressivity than a standard transformer. The delta rule underlying Delta Net
has inspired a range of further developments, including
RWKV-7 \citep{peng2025rwkv} and PaTH attention \citep{yang2025path}, that also
achieve state tracking beyond the capability of ordinary transformers while demonstrating
competitive language modeling at scale.
Similarly, other new forms of attention are being developed with
sequential dependencies that promise to be more powerful in stateful tasks 
\citep{beltagy2020longformer,Fagnou2024,lin2025forgetting,leviathan2025selective,fagnou2026},
including the 
notion of gated linear attention \citep{Yang2024gatedlinearattention} and gated
Delta Net \citep{yang2025gated} that, when mixed with
standard transformer blocks is more powerful than either, both in
theory and practice \citep{merrill2026olmohybrid}.

\subsection{Approximating state tracking in feedforward transformers}
Rather than incorporating recurrence, feedforward transformers can be steered 
to better approximate state tracking through specialized training objectives and 
structural priors \citep{hu2025,teoh2025,huang2026semantictubepredictionbeating}.
The promise is that such biases will encourage models to bolster their
native lookback abilities. However, we hope that future research will take
into account structured, compositional state representations.

\subsection{Coarse recurrence}
Introducing recurrence at a coarser granularity than individual tokens can mitigate 
the computational burden of token-by-token state updates. In the past, block-recurrent
models have operated on this principle by compressing and passing memory forward in fixed-
length chunks. A promising approach is to consider linguistic structure in identifying
the chunks: \citet{borazjanizadeh2026} have modeled language as a sequence of discrete `thoughts' by chunking at the sentence level.

\subsection{Leveraging representational alignment}
Variable-depth models dynamically select the number of iterations of a layer or
the number of times a range of layers repeats. The fact that this approach succeeds
with only fine tuning---and sometimes with no training whatsoever---suggests that 
the model is well predisposed to communication of representations across layers,
due to the alignment resulting from the residual connections. Similarly,
canon layers \citep{allenzhu2025physicslanguagemodels1} leverage alignment of
representations from one input step to the next. We speculate that there are
additional means of leveraging this alignment.

\subsection{Efficient training of recurrence}
Any architecture which is capable of indefinite, arbitrary state propagation 
requires autoregressive processing. Autoregressive pretraining is 
computationally inefficient and limits parallelization \citep{chevalier2023adaptinglanguagemodelscompress}.  One potential solution is
a multi-stage training scheme in which initial pretraining relies entirely on
standard, parallelizable feedforward transformer architectures.
Recurrent mechanisms are then introduced only at a later training stage. 
To ensure training efficiency during these subsequent recurrent stages, 
optimization techniques such as truncated gradient methods 
and---for training attractor dynamics---recurrent backpropagation 
\citep{almeida1987learning,pineda1987generalization,liao2018}.
Methods have also been proposed to increase \emph{arithmetic intensity} to obtain scaling
that is near linear time in context length versus quadratic for a naive implementation
\citep{oncescu2026}.

\section{Conclusions}
Although the transformer’s feedforward design has expanded the limits of context-based
retrieval, its topological structure remains fundamentally at odds with the 
iterative nature of state tracking. As we have argued, the current reliance on 
explicit natural-language-like ``thought'' to bypass depth limitations is an inefficient workaround for a structural deficiency. By transitioning toward implicit, recurrent activation dynamics, we can move beyond these depth-limited constraints to attain 
robust long-term coherence and multihop inference.

The taxonomy and research directions proposed in this article provide a roadmap 
for improving sequential inference dependencies without sacrificing the 
foundational strengths of modern models. Ultimately, bridging the gap between 
the transformer's parallel efficiency and the brain's inherent dynamical nature 
is essential. The next generation of foundation models must do more than simply 
re-scan the past; they must maintain a fluid, evolving representation of reality 
that persists across the many time scales required for temporally extended cognition.
\begin{ack}
Many thanks to Sunny Sanyal, Kazuki Irie, Kevin Murphy, Jay McClelland, and Chris Williams for helpful feedback on earlier drafts of the manuscript.
\end{ack}
\newpage
\bibliography{topological_trouble}

@misc{raposo2024,
      title={Mixture-of-Depths: Dynamically allocating compute in transformer-based language models}, 
      author={David Raposo and Sam Ritter and Blake Richards and Timothy Lillicrap and Peter Conway Humphreys and Adam Santoro},
      year={2024},
      eprint={2404.02258},
      archivePrefix={arXiv},
      primaryClass={cs.LG},
      url={https://arxiv.org/abs/2404.02258}, 
      note={arXiv:2404.02258 [cs.LG]},
}

@misc{chen2025,
      title={Inner Thinking Transformer: Leveraging Dynamic Depth Scaling to Foster Adaptive Internal Thinking}, 
      author={Yilong Chen and Junyuan Shang and Zhenyu Zhang and Yanxi Xie and Jiawei Sheng and Tingwen Liu and Shuohuan Wang and Yu Sun and Hua Wu and Haifeng Wang},
      year={2025},
      eprint={2502.13842},
      archivePrefix={arXiv},
      primaryClass={cs.CL},
      note={arXiv:2502.13842 [cs.CL]}, 
}

@inproceedings{Mozer1992,
  author    = {Mozer, Michael C.},
  title     = {The induction of multiscale temporal structure},
  booktitle = {Advances in Neural Information Processing Systems 4},
  editor    = {Moody, John E. and Hanson, Stephen J. and Lippmann, Richard P.},
  year      = {1992},
  pages     = {275--282},
  publisher = {Morgan Kaufmann},
  address   = {San Mateo, CA}
}

@inproceedings{Vaswanietal2017,
  author    = {Vaswani, Ashish and Shazeer, Noam and Parmar, Niki and Uszkoreit, Jakob and Jones, Llion and Gomez, Aidan N. and Kaiser, {\L}ukasz and Polosukhin, Illia},
  title     = {Attention Is All You Need},
  booktitle = {Advances in Neural Information Processing Systems},
  editor = {I. Guyon and U. Von Luxburg and S. Bengio and H. Wallach and R. Fergus and S. Vishwanathan and R. Garnett},
  year      = {2017},
  volume    = {30},
  pages     = {5998--6008},
  publisher = {Curran Associates, Inc.},
  url       = {https://papers.nips.cc/paper/7181-attention-is-all-you-need}
}

@misc{allenzhu2025,
      title={Physics of Language Models: Part 4.1, Architecture Design and the Magic of Canon Layers}, 
      author={Zeyuan Allen-Zhu},
      year={2025},
      eprint={2512.17351},
      archivePrefix={arXiv},
      primaryClass={cs.CL},
      url={https://arxiv.org/abs/2512.17351}, 
      note={arXiv:2512.17351 [cs.CL]}, 
}

@misc{allenzhu2025physicslanguagemodels1,
      title={Physics of Language Models: Part 1, Learning Hierarchical Language Structures}, 
      author={Zeyuan Allen-Zhu and Yuanzhi Li},
      year={2025},
      eprint={2305.13673},
      archivePrefix={arXiv},
      primaryClass={cs.CL},
      url={https://arxiv.org/abs/2305.13673}, 
      note={arXiv:2305.13673 [cs.CL]}, 
}

@article{hochreiter1998,
  title={The vanishing gradient problem during learning recurrent neural nets and problem solutions},
  author={Hochreiter, Sepp},
  journal={International Journal of Uncertainty, Fuzziness and Knowledge-Based Systems},
  volume={6},
  number={02},
  pages={107--116},
  year={1998},
  publisher={World Scientific}
}

@misc{hochreiter2001,
  title={Gradient flow in recurrent nets: the difficulty of learning long-term dependencies},
  author={Hochreiter, Sepp and Bengio, Yoshua and Frasconi, Paolo and Schmidhuber, J{\"u}rgen},
  year={2001},
  editor={John F. Kolen and  Stefan C. Kremer},
  publisher={A field guide to dynamical recurrent neural networks. IEEE Press In}
}

@misc{laban2025,
      title={LLMs Get Lost In Multi-Turn Conversation}, 
      author={Philippe Laban and Hiroaki Hayashi and Yingbo Zhou and Jennifer Neville},
      year={2025},
      eprint={2505.06120},
      archivePrefix={arXiv},
      primaryClass={cs.CL},
      url={https://arxiv.org/abs/2505.06120}, 
      note={arXiv:2505.06120 [cs.CL]},
}

@inproceedings{lepori2025,
    title = "Racing Thoughts: Explaining Contextualization Errors in Large Language Models",
    author = "Lepori, Michael A.  and
      Mozer, Michael Curtis  and
      Ghandeharioun, Asma",
    editor = "Chiruzzo, Luis  and
      Ritter, Alan  and
      Wang, Lu",
    booktitle = "Proceedings of the 2025 Conference of the Nations of the Americas Chapter of the Association for Computational Linguistics: Human Language Technologies (Volume 1: Long Papers)",
    month = apr,
    year = "2025",
    address = "Albuquerque, New Mexico",
    publisher = "Association for Computational Linguistics",
    url = "https://aclanthology.org/2025.naacl-long.155/",
    doi = "10.18653/v1/2025.naacl-long.155",
    pages = "3020--3036",
    ISBN = "979-8-89176-189-6"
}

@article{Vul2014,
  author = {Vul, Edward and Goodman, Noah and Griffiths, Thomas L. and Tenenbaum, Joshua B.},
  title = {One and Done? {O}ptimal Decisions From Very Few Samples},
  journal = {Cognitive Science},
  year = {2014},
  volume = {38},
  number = {4},
  pages = {599--637},
  doi = {10.1111/cogs.12101},
}

@book{johnson1983,
  title={Mental Models: Towards a Cognitive Science of Language, Inference, and Consciousness},
  author={Johnson-Laird, P.N.},
  isbn={9780674568822},
  lccn={83004333},
  series={Cognitive science series},
  url={https://books.google.com/books?id=FS3zSKAfLGMC},
  year={1983},
  publisher={Harvard University Press}
}

@article{Tversky1971,
  author = {Tversky, Amos and Kahneman, Daniel},
  title = {Belief in the Law of Small Numbers},
  journal = {Psychological Bulletin},
  year = {1971},
  volume = {76},
  number = {2},
  pages = {105--110},
  doi = {10.1037/h0031322},
  publisher = {American Psychological Association}
}

@misc{khatua2026,
      title={CooperBench: Why Coding Agents Cannot be Your Teammates Yet}, 
      author={Arpandeep Khatua and Hao Zhu and Peter Tran and Arya Prabhudesai and Frederic Sadrieh and Johann K. Lieberwirth and Xinkai Yu and Yicheng Fu and Michael J. Ryan and Jiaxin Pei and Diyi Yang},
      year={2026},
      eprint={2601.13295},
      archivePrefix={arXiv},
      primaryClass={cs.LG},
      url={https://arxiv.org/abs/2601.13295}, 
}

@misc{davidson2025,
      title={The Collaboration Gap}, 
      author={Tim R. Davidson and Adam Fourney and Saleema Amershi and Robert West and Eric Horvitz and Ece Kamar},
      year={2025},
      eprint={2511.02687},
      archivePrefix={arXiv},
      primaryClass={cs.AI},
      url={https://arxiv.org/abs/2511.02687}, 
      note={arXiv:2511.02687 [cs.AI]}, 
}

@misc{merrill2025,
      title={A Little Depth Goes a Long Way: The Expressive Power of Log-Depth Transformers}, 
      author={William Merrill and Ashish Sabharwal},
      year={2025},
      eprint={2503.03961},
      archivePrefix={arXiv},
      primaryClass={cs.LG},
      url={https://arxiv.org/abs/2503.03961}, 
      note={arXiv:2503.03961 [cs.LG]},
}

@misc{merrill2025illusion,
      title={The Illusion of State in State-Space Models}, 
      author={William Merrill and Jackson Petty and Ashish Sabharwal},
      year={2025},
      eprint={2404.08819},
      archivePrefix={arXiv},
      primaryClass={cs.LG},
      url={https://arxiv.org/abs/2404.08819}, 
      note={arXiv:2404.08819 [cs.LG]},
}

@inproceedings{
peng2025rwkv,
title={{RWKV}-7 ''Goose'' with Expressive Dynamic State Evolution},
author={Bo Peng and Ruichong Zhang and Daniel Goldstein and Eric Alcaide and Xingjian Du and Haowen Hou and Jiaju Lin and Jiaxing Liu and Janna Lu and William Merrill and Guangyu Song and Kaifeng Tan and Saiteja Utpala and Nathan Wilce and Johan S. Wind and Tianyi Wu and Daniel Wuttke and Christian Zhou-Zheng},
booktitle={Second Conference on Language Modeling},
year={2025},
url={https://openreview.net/forum?id=ayB1PACN5j}
}

@article{Lindsey2025,
  author  = {Lindsey, Jack and Gurnee, Wes and Ameisen, Emmanuel and Chen, Becca and Pearce, Adam and Turner, Nelson L. and Citro, Chris and others},
  title   = {On the Biology of a Large Language Model},
  journal = {Transformer Circuits Thread},
  year    = {2025},
  url     = {https://transformer-circuits.pub/2025/attribution-graphs/biology.html}
}

@inproceedings{
li2025,
title={(How) Do Language Models Track State?},
author={Belinda Z. Li and Zifan Carl Guo and Jacob Andreas},
booktitle={Forty-second International Conference on Machine Learning},
year={2025},
url={https://openreview.net/forum?id=8SXosAVIFH}
}

@inproceedings{
hu2025,
title={The Belief State Transformer},
author={Edward S. Hu and Kwangjun Ahn and Qinghua Liu and Haoran Xu and Manan Tomar and Ada Langford and Dinesh Jayaraman and Alex Lamb and John Langford},
booktitle={The Thirteenth International Conference on Learning Representations},
year={2025},
url={https://openreview.net/forum?id=ThRMTCgpvo}
}

@misc{piotrowski2025,
      title={Constrained belief updates explain geometric structures in transformer representations}, 
      author={Mateusz Piotrowski and Paul M. Riechers and Daniel Filan and Adam S. Shai},
      year={2025},
      eprint={2502.01954},
      archivePrefix={arXiv},
      primaryClass={cs.LG},
      url={https://arxiv.org/abs/2502.01954}, 
      note={arXiv:2502.01954 [cs.LG]}, 
}

@inproceedings{
prakash2026,
title={Language Models Use Lookbacks to Track Beliefs},
author={Nikhil Prakash and Natalie Shapira and Arnab Sen Sharma and Christoph Riedl and Yonatan Belinkov and Tamar Rott Shaham and David Bau and Atticus Geiger},
booktitle={The Fourteenth International Conference on Learning Representations},
year={2026},
url={https://openreview.net/forum?id=6gO6KTRMpG}
}

@inproceedings{
shai2024,
title={Transformers Represent Belief State Geometry in their Residual Stream},
author={Adam Shai and Paul M. Riechers and Lucas Teixeira and Alexander Gietelink Oldenziel and Sarah Marzen},
booktitle={The Thirty-eighth Annual Conference on Neural Information Processing Systems},
year={2024},
url={https://openreview.net/forum?id=YIB7REL8UC}
}

@misc{teoh2025,
      title={Next-Latent Prediction Transformers Learn Compact World Models}, 
      author={Jayden Teoh and Manan Tomar and Kwangjun Ahn and Edward S. Hu and Pratyusha Sharma and Riashat Islam and Alex Lamb and John Langford},
      year={2025},
      eprint={2511.05963},
      archivePrefix={arXiv},
      primaryClass={cs.LG},
      url={https://arxiv.org/abs/2511.05963}, 
}

@inproceedings{ghandeharioun2024,
author = {Ghandeharioun, Asma and Caciularu, Avi and Pearce, Adam and Dixon, Lucas and Geva, Mor},
title = {Patchscopes: a unifying framework for inspecting hidden representations of language models},
year = {2024},
publisher = {JMLR.org},
booktitle = {Proceedings of the 41st International Conference on Machine Learning},
articleno = {620},
numpages = {25},
location = {Vienna, Austria},
series = {ICML'24}
}

@article{rumelhart1986,
  author = {Rumelhart, David E and Hinton, Geoffrey E and Williams, Ronald J},
  journal = {nature},
  number = 6088,
  pages = {533--536},
  title = {Learning representations by back-propagating errors},
  volume = 323,
  year = 1986
}

@article{merrill-sabharwal-2023-parallelism,
    title = "The Parallelism Tradeoff: Limitations of Log-Precision Transformers",
    author = "Merrill, William  and
      Sabharwal, Ashish",
    journal = "Transactions of the Association for Computational Linguistics",
    volume = "11",
    year = "2023",
    address = "Cambridge, MA",
    publisher = "MIT Press",
    url = "https://aclanthology.org/2023.tacl-1.31/",
    doi = "10.1162/tacl_a_00562",
    pages = "531--545"
}

@article{Strobl2024,
    author = {Strobl, Lena and Merrill, William and Weiss, Gail and Chiang, David and Angluin, Dana},
    title = {What Formal Languages Can Transformers Express? A Survey},
    journal = {Transactions of the Association for Computational Linguistics},
    volume = {12},
    pages = {543-561},
    year = {2024},
    month = {05},
    issn = {2307-387X},
    doi = {10.1162/tacl_a_00663},
    url = {https://doi.org/10.1162/tacl_a_00663},
    eprint = {https://direct.mit.edu/tacl/article-pdf/doi/10.1162/tacl_a_00663/2370911/tacl_a_00663.pdf},
}

@inproceedings{biran2024,
    title = "Hopping Too Late: Exploring the Limitations of Large Language Models on Multi-Hop Queries",
    author = "Biran, Eden  and
      Gottesman, Daniela  and
      Yang, Sohee  and
      Geva, Mor  and
      Globerson, Amir",
    editor = "Al-Onaizan, Yaser  and
      Bansal, Mohit  and
      Chen, Yun-Nung",
    booktitle = "Proceedings of the 2024 Conference on Empirical Methods in Natural Language Processing",
    month = nov,
    year = "2024",
    address = "Miami, Florida, USA",
    publisher = "Association for Computational Linguistics",
    url = "https://aclanthology.org/2024.emnlp-main.781/",
    doi = "10.18653/v1/2024.emnlp-main.781",
    pages = "14113--14130"
}

@inproceedings{
venhoff2025,
title={Too Late to Recall: Explaining the Two-Hop Problem in Multimodal Knowledge Retrieval},
author={Constantin Venhoff and Ashkan Khakzar and Sonia Joseph and Philip Torr and Neel Nanda},
booktitle={The Thirty-ninth Annual Conference on Neural Information Processing Systems},
year={2025},
url={https://openreview.net/forum?id=qeL8fi8GS7}
}

@inproceedings{
sawyer2025,
title={Exploring exploration with foundation agents in interactive environments},
author={Daniel P. Sawyer and Nan Rosemary Ke and Hubert Soyer and Martin Engelcke and John Reid and David P Reichert and Drew A. Hudson and Alexander Lerchner and Danilo Jimenez Rezende and Timothy P Lillicrap and Michael Curtis Mozer and Jane X Wang},
booktitle={NeurIPS 2025 Workshop on Embodied World Models for Decision Making},
year={2025},
url={https://openreview.net/forum?id=ay2d66HWO2}
}

@misc{mcleish2025,
      title={Teaching Pretrained Language Models to Think Deeper with Retrofitted Recurrence}, 
      author={Sean McLeish and Ang Li and John Kirchenbauer and Dayal Singh Kalra and Brian R. Bartoldson and Bhavya Kailkhura and Avi Schwarzschild and Jonas Geiping and Tom Goldstein and Micah Goldblum},
      year={2025},
      eprint={2511.07384},
      archivePrefix={arXiv},
      primaryClass={cs.CL},
      url={https://arxiv.org/abs/2511.07384}, 
}

@misc{koishekenov2025,
      title={Encode, Think, Decode: Scaling test-time reasoning with recursive latent thoughts}, 
      author={Yeskendir Koishekenov and Aldo Lipani and Nicola Cancedda},
      year={2025},
      eprint={2510.07358},
      archivePrefix={arXiv},
      primaryClass={cs.LG},
      url={https://arxiv.org/abs/2510.07358}, 
      note={arXiv:2510.07358 [cs.LG]},
}

@misc{li2025skiplayerloopit,
      title={Skip a Layer or Loop it? Test-Time Depth Adaptation of Pretrained LLMs}, 
      author={Ziyue Li and Yang Li and Tianyi Zhou},
      year={2025},
      eprint={2507.07996},
      archivePrefix={arXiv},
      primaryClass={cs.LG},
      url={https://arxiv.org/abs/2507.07996}, 
      note={arXiv:2507.07996 [cs.LG]}
}

@inproceedings{
geiping2025,
title={Scaling up Test-Time Compute with Latent Reasoning: A Recurrent Depth Approach},
author={Jonas Geiping and Sean Michael McLeish and Neel Jain and John Kirchenbauer and Siddharth Singh and Brian R. Bartoldson and Bhavya Kailkhura and Abhinav Bhatele and Tom Goldstein},
booktitle={The Thirty-ninth Annual Conference on Neural Information Processing Systems},
year={2025},
url={https://openreview.net/forum?id=S3GhJooWIC}
}

@misc{rodkin2025,
      title={Beyond Memorization: Extending Reasoning Depth with Recurrence, Memory and Test-Time Compute Scaling}, 
      author={Ivan Rodkin and Daniil Orel and Konstantin Smirnov and Arman Bolatov and Bilal Elbouardi and Besher Hassan and Yuri Kuratov and Aydar Bulatov and Preslav Nakov and Timothy Baldwin and Artem Shelmanov and Mikhail Burtsev},
      year={2025},
      eprint={2508.16745},
      archivePrefix={arXiv},
      primaryClass={cs.LG},
      url={https://arxiv.org/abs/2508.16745}, 
      note={arXiv:2508.16745 [cs.LG]},
}

@inproceedings{
bae2025,
title={Mixture-of-Recursions: Learning Dynamic Recursive Depths for Adaptive Token-Level Computation},
author={Sangmin Bae and Yujin Kim and Reza Bayat and Sungnyun Kim and Jiyoun Ha and Tal Schuster and Adam Fisch and Hrayr Harutyunyan and Ziwei Ji and Aaron Courville and Se-Young Yun},
booktitle={The Thirty-ninth Annual Conference on Neural Information Processing Systems},
year={2025},
url={https://openreview.net/forum?id=QuqsEIVWIG}
}

@inproceedings{
yang2024,
title={Looped Transformers are Better at Learning Learning Algorithms},
author={Liu Yang and Kangwook Lee and Robert D Nowak and Dimitris Papailiopoulos},
booktitle={The Twelfth International Conference on Learning Representations},
year={2024},
url={https://openreview.net/forum?id=HHbRxoDTxE}
}

@misc{zhu2025,
      title={Scaling Latent Reasoning via Looped Language Models}, 
      author={Rui-Jie Zhu and Zixuan Wang and Kai Hua and Tianyu Zhang and Ziniu Li and Haoran Que and Boyi Wei and Zixin Wen and Fan Yin and He Xing and Lu Li and Jiajun Shi and Kaijing Ma and Shanda Li and Taylor Kergan and Andrew Smith and Xingwei Qu and Mude Hui and Bohong Wu and Qiyang Min and Hongzhi Huang and Xun Zhou and Wei Ye and Jiaheng Liu and Jian Yang and Yunfeng Shi and Chenghua Lin and Enduo Zhao and Tianle Cai and Ge Zhang and Wenhao Huang and Yoshua Bengio and Jason Eshraghian},
      year={2025},
      eprint={2510.25741},
      archivePrefix={arXiv},
      primaryClass={cs.CL},
      url={https://arxiv.org/abs/2510.25741}, 
      note={arXiv:2510.25741 [cs.LG]},
}

@inproceedings{
saunshi2025,
title={Reasoning with Latent Thoughts: On the Power of Looped Transformers},
author={Nikunj Saunshi and Nishanth Dikkala and Zhiyuan Li and Sanjiv Kumar and Sashank J. Reddi},
booktitle={The Thirteenth International Conference on Learning Representations},
year={2025},
url={https://openreview.net/forum?id=din0lGfZFd}
}

@inproceedings{hutchins2022,
 author = {Hutchins, DeLesley and Schlag, Imanol and Wu, Yuhuai and Dyer, Ethan and Neyshabur, Behnam},
 booktitle = {Advances in Neural Information Processing Systems},
 editor = {S. Koyejo and S. Mohamed and A. Agarwal and D. Belgrave and K. Cho and A. Oh},
 pages = {33248--33261},
 publisher = {Curran Associates, Inc.},
 title = {Block-Recurrent Transformers},
 url = {https://proceedings.neurips.cc/paper_files/paper/2022/file/d6e0bbb9fc3f4c10950052ec2359355c-Paper-Conference.pdf},
 volume = {35},
 year = {2022}
}

@misc{borazjanizadeh2026,
      title={Modeling Language as a Sequence of Thoughts}, 
      author={Nasim Borazjanizadeh and James McClelland},
      year={2025},
      eprint={2512.25026},
      archivePrefix={arXiv},
      primaryClass={cs.CL},
      url={https://arxiv.org/abs/2512.25026}, 
      note={arXiv:2512.25026 [cs.CL]}, 
}

@inproceedings{
gu2024,
title={Mamba: Linear-Time Sequence Modeling with Selective State Spaces},
author={Albert Gu and Tri Dao},
booktitle={First Conference on Language Modeling},
year={2024},
url={https://openreview.net/forum?id=tEYskw1VY2}
}

@misc{jolicoeurmartineau2025,
      title={Less is More: Recursive Reasoning with Tiny Networks}, 
      author={Alexia Jolicoeur-Martineau},
      year={2025},
      eprint={2510.04871},
      archivePrefix={arXiv},
      primaryClass={cs.LG},
      url={https://arxiv.org/abs/2510.04871}, 
      note={arXiv:2510.04871 [cs.LG]}
}

@inproceedings{giannou2023looped,
  title={Looped transformers as programmable computers},
  author={Giannou, Angeliki and Rajput, Shashank and Sohn, Jy-yong and Lee, Kangwook and Lee, Jason D and Papailiopoulos, Dimitris},
  booktitle={International Conference on Machine Learning},
  pages={11398--11442},
  year={2023},
  organization={PMLR}
}

@inproceedings{
dehghani2018universal,
title={Universal Transformers},
author={Mostafa Dehghani and Stephan Gouws and Oriol Vinyals and Jakob Uszkoreit and Lukasz Kaiser},
booktitle={International Conference on Learning Representations},
year={2019},
url={https://openreview.net/forum?id=HyzdRiR9Y7},
}

@inproceedings{bulatov2022,
 author = {Bulatov, Aydar and Kuratov, Yury and Burtsev, Mikhail},
 booktitle = {Advances in Neural Information Processing Systems},
 editor = {S. Koyejo and S. Mohamed and A. Agarwal and D. Belgrave and K. Cho and A. Oh},
 pages = {11079--11091},
 publisher = {Curran Associates, Inc.},
 title = {Recurrent Memory Transformer},
 url = {https://proceedings.neurips.cc/paper_files/paper/2022/file/47e288629a6996a17ce50b90a056a0e1-Paper-Conference.pdf},
 volume = {35},
 year = {2022}
}

@inproceedings{
zeng2026ponderlm,
title={Ponder{LM}: Pretraining Language Models to Ponder in Continuous Space},
author={Boyi Zeng and Shixiang Song and Siyuan Huang and Yixuan Wang and He Li and Ziwei He and Xinbing Wang and Zhiyu li and Zhouhan Lin},
booktitle={The Fourteenth International Conference on Learning Representations},
year={2026},
url={https://openreview.net/forum?id=UrM4MNRYZm}
}

@misc{nowak2024,
      title={Towards Optimal Adapter Placement for Efficient Transfer Learning}, 
      author={Aleksandra I. Nowak and Otniel-Bogdan Mercea and Anurag Arnab and Jonas Pfeiffer and Yann Dauphin and Utku Evci},
      year={2024},
      eprint={2410.15858},
      archivePrefix={arXiv},
      primaryClass={cs.LG},
      url={https://arxiv.org/abs/2410.15858}, 
      note={arXiv:2410.15858 [cs.LG]},
}

@inproceedings{yu2025backattn,
    title = "Back Attention: Understanding and Enhancing Multi-Hop Reasoning in Large Language Models",
    author = "Yu, Zeping  and
      Belinkov, Yonatan  and
      Ananiadou, Sophia",
    editor = "Christodoulopoulos, Christos  and
      Chakraborty, Tanmoy  and
      Rose, Carolyn  and
      Peng, Violet",
    booktitle = "Proceedings of the 2025 Conference on Empirical Methods in Natural Language Processing",
    month = nov,
    year = "2025",
    address = "Suzhou, China",
    publisher = "Association for Computational Linguistics",
    url = "https://aclanthology.org/2025.emnlp-main.567/",
    doi = "10.18653/v1/2025.emnlp-main.567",
    pages = "11257--11272",
    ISBN = "979-8-89176-332-6"
}

@inproceedings{
chen2025melodi,
title={{MELODI}: Exploring Memory Compression for Long Contexts},
author={Yinpeng Chen and DeLesley Hutchins and Aren Jansen and Andrey Zhmoginov and David Racz and Jesper Sparre Andersen},
booktitle={The Thirteenth International Conference on Learning Representations},
year={2025},
url={https://openreview.net/forum?id=TvGPP8i18S}
}

@inproceedings{
yang2025path,
title={Pa{TH} Attention: Position Encoding via Accumulating Householder Transformations},
author={Songlin Yang and Yikang Shen and Kaiyue Wen and Shawn Tan and Mayank Mishra and Liliang Ren and Rameswar Panda and Yoon Kim},
booktitle={The Thirty-ninth Annual Conference on Neural Information Processing Systems},
year={2025},
url={https://openreview.net/forum?id=ZBlHEeSvKd}
}

@misc{galashov2025,
      title={Catch Your Breath: Adaptive Computation for Self-Paced Sequence Production}, 
      author={Alexandre Galashov and Matt Jones and Rosemary Ke and Yuan Cao and Vaishnavh Nagarajan and Michael C. Mozer},
      year={2025},
      eprint={2510.13879},
      archivePrefix={arXiv},
      primaryClass={cs.CL},
      url={https://arxiv.org/abs/2510.13879}, 
      note={arXiv:2510.13879 [cs.CL]}, 
}

@misc{jabri2023,
      title={Scalable Adaptive Computation for Iterative Generation}, 
      author={Allan Jabri and David Fleet and Ting Chen},
      year={2023},
      eprint={2212.11972},
      archivePrefix={arXiv},
      primaryClass={cs.LG},
      url={https://arxiv.org/abs/2212.11972}, 
      note={arXiv:2212.11972 [cs.LG]}, 
}

@inproceedings{Katharopoulos2020Transformers,
  author    = {Katharopoulos, Angelos and Vyas, Apoorv and Pappas, Nikolaos and Fleuret, François},
  title     = {Transformers are {RNN}s: Fast Autoregressive Transformers with Linear Attention},
  booktitle = {Proceedings of the 37th International Conference on Machine Learning},
  series    = {Proceedings of Machine Learning Research},
  volume    = {119},
  pages     = {5156--5165},
  year      = {2020},
  publisher = {PMLR}
}

@misc{fan2021,
      title={Addressing Some Limitations of Transformers with Feedback Memory}, 
      author={Angela Fan and Thibaut Lavril and Edouard Grave and Armand Joulin and Sainbayar Sukhbaatar},
      year={2021},
      eprint={2002.09402},
      archivePrefix={arXiv},
      primaryClass={cs.LG},
      url={https://arxiv.org/abs/2002.09402}, 
      note={arXiv:2002.09402 [cs.CL]}, 
}

@inproceedings{
hao2025coconut,
title={Training Large Language Models to Reason in a Continuous Latent Space},
author={Shibo Hao and Sainbayar Sukhbaatar and DiJia Su and Xian Li and Zhiting Hu and Jason E Weston and Yuandong Tian},
booktitle={Second Conference on Language Modeling},
year={2025},
url={https://openreview.net/forum?id=Itxz7S4Ip3}
}

@misc{huang2026semantictubepredictionbeating,
      title={Semantic Tube Prediction: Beating {LLM} Data Efficiency with {JEPA}}, 
      author={Hai Huang and Yann LeCun and Randall Balestriero},
      year={2026},
      eprint={2602.22617},
      archivePrefix={arXiv},
      primaryClass={cs.LG},
      url={https://arxiv.org/abs/2602.22617}, 
      note={arXiv:2602.22617 [cs.LG]},
}

@misc{liu2026serialscalinghypothesis,
      title={The Serial Scaling Hypothesis}, 
      author={Yuxi Liu and Konpat Preechakul and Kananart Kuwaranancharoen and Yutong Bai},
      year={2026},
      eprint={2507.12549},
      archivePrefix={arXiv},
      primaryClass={cs.LG},
      url={https://arxiv.org/abs/2507.12549}, 
}

@inproceedings{
li2024chain,
title={Chain of Thought Empowers Transformers to Solve Inherently Serial Problems},
author={Zhiyuan Li and Hong Liu and Denny Zhou and Tengyu Ma},
booktitle={The Twelfth International Conference on Learning Representations},
year={2024},
url={https://openreview.net/forum?id=3EWTEy9MTM}
}

@inproceedings{
merrill2024cot,
title={The Expressive Power of Transformers with Chain of Thought},
author={William Merrill and Ashish Sabharwal},
booktitle={The Twelfth International Conference on Learning Representations},
year={2024},
url={https://openreview.net/forum?id=NjNGlPh8Wh}
}

@misc{ng2026rys,
  title        = {{LLM} Neuroanatomy: How {I} Topped the {LLM} Leaderboard Without Changing a Single Weight},
  author       = {Ng, David Noel},
  year         = {2026},
  month        = {March},
  howpublished = {\url{https://dnhkng.github.io/posts/rys/}}
}

@article{mozer1991induction,
  title={Induction of multiscale temporal structure},
  author={Mozer, Michael C},
  journal={Advances in neural information processing systems},
  volume={4},
  year={1991}
}

@article{Bengio1994,
  author={Bengio, Y. and Simard, P. and Frasconi, P.},
  journal={IEEE Transactions on Neural Networks}, 
  title={Learning long-term dependencies with gradient descent is difficult}, 
  year={1994},
  volume={5},
  number={2},
  pages={157-166},
  doi={10.1109/72.279181}}

@InProceedings{liao2018,
  title = 	 {Reviving and Improving Recurrent Back-Propagation},
  author =       {Liao, Renjie and Xiong, Yuwen and Fetaya, Ethan and Zhang, Lisa and Yoon, KiJung and Pitkow, Xaq and Urtasun, Raquel and Zemel, Richard},
  booktitle = 	 {Proceedings of the 35th International Conference on Machine Learning},
  pages = 	 {3082--3091},
  year = 	 {2018},
  editor = 	 {Dy, Jennifer and Krause, Andreas},
  volume = 	 {80},
  series = 	 {Proceedings of Machine Learning Research},
  month = 	 {10--15 Jul},
  publisher =    {PMLR},
  url = 	 {https://proceedings.mlr.press/v80/liao18c.html}
}

@article{pineda1987generalization,
  author  = {Pineda, Fernando},
  title   = {Generalization of Back-Propagation to Recurrent Neural Networks},
  journal = {Physical Review Letters},
  volume  = {59},
  number  = {19},
  pages   = {2229--2232},
  year    = {1987},
  month   = {nov},
  doi     = {10.1103/PhysRevLett.59.2229},
}

@inproceedings{almeida1987learning,
  author    = {Almeida, Luis B.},
  title     = {A Learning Rule for Asynchronous Perceptrons with Feedback in a Combinatorial Environment},
  booktitle = {Proceedings of the IEEE First International Conference on Neural Networks},
  volume    = {2},
  pages     = {609--618},
  year      = {1987},
  month     = {jun},
  address   = {San Diego, CA, USA},
  publisher = {IEEE}
}

@inproceedings{
jeddi2026loopformer,
title={LoopFormer: Elastic-Depth Looped Transformers for Latent Reasoning via Shortcut Modulation},
author={Ahmadreza Jeddi and Marco Ciccone and Babak Taati},
booktitle={The Fourteenth International Conference on Learning Representations},
year={2026},
url={https://openreview.net/forum?id=RzYXb5YWBs}
}

@misc{grant2025,
      title={Emergent Symbol-like Number Variables in Artificial Neural Networks}, 
      author={Satchel Grant and Noah D. Goodman and James L. McClelland},
      year={2025},
      eprint={2501.06141},
      archivePrefix={arXiv},
      primaryClass={cs.LG},
      url={https://arxiv.org/abs/2501.06141}, 
}

@inproceedings{
lin2025forgetting,
title={Forgetting Transformer: Softmax Attention with a Forget Gate},
author={Zhixuan Lin and Evgenii Nikishin and Xu He and Aaron Courville},
booktitle={The Thirteenth International Conference on Learning Representations},
year={2025},
url={https://openreview.net/forum?id=q2Lnyegkr8}
}

@inproceedings{
leviathan2025selective,
title={Selective Attention Improves Transformer},
author={Yaniv Leviathan and Matan Kalman and Yossi Matias},
booktitle={The Thirteenth International Conference on Learning Representations},
year={2025},
url={https://openreview.net/forum?id=v0FzmPCd1e}
}

@inproceedings{
alabdulmohsin2025recursive,
title={Recursive Inference Scaling: A Winning Path to Scalable Inference in Language and Multimodal Systems},
author={Ibrahim Alabdulmohsin and Xiaohua Zhai},
booktitle={The Thirty-ninth Annual Conference on Neural Information Processing Systems},
year={2025},
url={https://openreview.net/forum?id=cLbGkINOLP}
}

@misc{chevalier2023adaptinglanguagemodelscompress,
      title={Adapting Language Models to Compress Contexts}, 
      author={Alexis Chevalier and Alexander Wettig and Anirudh Ajith and Danqi Chen},
      year={2023},
      eprint={2305.14788},
      archivePrefix={arXiv},
      primaryClass={cs.CL},
      url={https://arxiv.org/abs/2305.14788}, 
      note={arXiv:2305.14788 [cs.CL]},
}

@inproceedings{yang2025gated,
title={Gated Delta Networks: Improving Mamba2 with Delta Rule},
author={Songlin Yang and Jan Kautz and Ali Hatamizadeh},
booktitle={The Thirteenth International Conference on Learning Representations},
year={2025},
url={https://openreview.net/forum?id=r8H7xhYPwz},
}

@techreport{merrill2026olmohybrid,
  title       = {Olmo Hybrid: From Theory to Practice and Back},
  author      = {Merrill, William and Li, Yanhong and Romero, Tyler and Svete, Anej and Costello, Caia and Dasigi, Pradeep and Groeneveld, Dirk and Heineman, David and Kuehl, Bailey and Lambert, Nathan and Li, Chuan and Lo, Kyle and Malik, Saumya and Matusz, DJ and Minixhofer, Benjamin and Morrison, Jacob and Soldaini, Luca and Timbers, Finbarr and Walsh, Pete and Smith, Noah A. and Hajishirzi, Hannaneh and Sabharwal, Ashish},
  institution = {Allen Institute for AI},
  year        = {2026},
  url         = {https://allenai.org/papers/olmo-hybrid},
  note        = {Technical Report}
}

@misc{grazzi2025,
      title={Unlocking State-Tracking in Linear RNNs Through Negative Eigenvalues}, 
      author={Riccardo Grazzi and Julien Siems and Arber Zela and Jörg K. H. Franke and Frank Hutter and Massimiliano Pontil},
      year={2025},
      eprint={2411.12537},
      archivePrefix={arXiv},
      primaryClass={cs.LG},
      url={https://arxiv.org/abs/2411.12537}, 
      note={arXiv:2411.12537 [cs.LG]},
}

@article{meng2022locating,
  title={Locating and editing factual associations in gpt},
  author={Meng, Kevin and Bau, David and Andonian, Alex and Belinkov, Yonatan},
  journal={Advances in neural information processing systems},
  volume={35},
  pages={17359--17372},
  year={2022}
}

@article{hochreiter1997long,
  title={Long short-term memory},
  author={Hochreiter, Sepp and Schmidhuber, J{\"u}rgen},
  journal={Neural computation},
  volume={9},
  number={8},
  pages={1735--1780},
  year={1997},
  publisher={MIT press}
}

@article{olsson2022context,
   title={In-context Learning and Induction Heads},
   author={Olsson, Catherine and Elhage, Nelson and Nanda, Neel and Joseph, Nicholas and DasSarma, Nova and Henighan, Tom and Mann, Ben and Askell, Amanda and Bai, Yuntao and Chen, Anna and Conerly, Tom and Drain, Dawn and Ganguli, Deep and Hatfield-Dodds, Zac and Hernandez, Danny and Johnston, Scott and Jones, Andy and Kernion, Jackson and Lovitt, Liane and Ndousse, Kamal and Amodei, Dario and Brown, Tom and Clark, Jack and Kaplan, Jared and McCandlish, Sam and Olah, Chris},
   year={2022},
   journal={Transformer Circuits Thread},
   note={https://transformer-circuits.pub/2022/in-context-learning-and-induction-heads/index.html}
}

@article{wei2022chain,
  title={Chain-of-thought prompting elicits reasoning in large language models},
  author={Wei, Jason and Wang, Xuezhi and Schuurmans, Dale and Bosma, Maarten and Xia, Fei and Chi, Ed and Le, Quoc V and Zhou, Denny and others},
  journal={Advances in neural information processing systems},
  volume={35},
  pages={24824--24837},
  year={2022}
}

@misc{dai2019transformerxlattentivelanguagemodels,
      title={Transformer-XL: Attentive Language Models Beyond a Fixed-Length Context}, 
      author={Zihang Dai and Zhilin Yang and Yiming Yang and Jaime Carbonell and Quoc V. Le and Ruslan Salakhutdinov},
      year={2019},
      eprint={1901.02860},
      archivePrefix={arXiv},
      primaryClass={cs.LG},
      url={https://arxiv.org/abs/1901.02860}, 
      note={arXiv:1901.02860 [cs.LG]},
}

@article{gloeckle2024better,
  title={Better and faster large language models via multi-token prediction},
  author={Gloeckle, Fabian and Idrissi, Badr Youbi and Rozi{\`e}re, Baptiste and Lopez-Paz, David and Synnaeve, Gabriel},
  journal={arXiv preprint arXiv:2404.19737},
  year={2024}
}

@article{teoh2025next,
  title={Next-Latent Prediction Transformers Learn Compact World Models},
  author={Teoh, Jayden and Tomar, Manan and Ahn, Kwangjun and Hu, Edward S and Sharma, Pratyusha and Islam, Riashat and Lamb, Alex and Langford, John},
  journal={arXiv preprint arXiv:2511.05963},
  year={2025}
}

@article{yang2024kvsharer,
  title={Kvsharer: Efficient inference via layer-wise dissimilar kv cache sharing},
  author={Yang, Yifei and Cao, Zouying and Chen, Qiguang and Qin, Libo and Yang, Dongjie and Zhao, Hai and Chen, Zhi},
  journal={arXiv preprint arXiv:2410.18517},
  year={2024}
}

@article{hu2022lora,
  title={Lora: Low-rank adaptation of large language models.},
  author={Hu, Edward J and Shen, Yelong and Wallis, Phillip and Allen-Zhu, Zeyuan and Li, Yuanzhi and Wang, Shean and Wang, Liang and Chen, Weizhu and others},
  journal={Iclr},
  volume={1},
  number={2},
  pages={3},
  year={2022}
}

@article{liu2022transformers,
  title={Transformers learn shortcuts to automata},
  author={Liu, Bingbin and Ash, Jordan T and Goel, Surbhi and Krishnamurthy, Akshay and Zhang, Cyril},
  journal={arXiv preprint arXiv:2210.10749},
  year={2022}
}

@article{beltagy2020longformer,
  title={Longformer: The long-document transformer},
  author={Beltagy, Iz and Peters, Matthew E and Cohan, Arman},
  journal={arXiv preprint arXiv:2004.05150},
  year={2020}
}

@article{kaelbling1998planning,
title = {Planning and acting in partially observable stochastic domains},
author = {Leslie Pack Kaelbling and Michael L. Littman and Anthony R. Cassandra},
journal = {Artificial Intelligence},
volume = {101},
number = {1},
pages = {99-134},
year = {1998},
issn = {0004-3702},
doi = {https://doi.org/10.1016/S0004-3702(98)00023-X},
url = {https://www.sciencedirect.com/science/article/pii/S000437029800023X},
}

@inproceedings{Chrisman1992,
author = {Chrisman, Lonnie},
title = {Reinforcement learning with perceptual aliasing: the perceptual distinctions approach},
year = {1992},
isbn = {0262510634},
publisher = {AAAI Press},
booktitle = {Proceedings of the Tenth National Conference on Artificial Intelligence},
pages = {183–188},
numpages = {6},
location = {San Jose, California},
series = {AAAI'92}
}

@inproceedings{
sun2025learning,
title={Learning to (Learn at Test Time): {RNN}s with Expressive Hidden States},
author={Yu Sun and Xinhao Li and Karan Dalal and Jiarui Xu and Arjun Vikram and Genghan Zhang and Yann Dubois and Xinlei Chen and Xiaolong Wang and Sanmi Koyejo and Tatsunori Hashimoto and Carlos Guestrin},
booktitle={Forty-second International Conference on Machine Learning},
year={2025},
url={https://openreview.net/forum?id=wXfuOj9C7L}
}

@article{schlag2021,
      title={Linear Transformers Are Secretly Fast Weight Programmers}, 
      author={Imanol Schlag and Kazuki Irie and Jürgen Schmidhuber},
      journal={arXiv preprint arXiv:2102.11174},
      year={2021},
}

@misc{oncescu2026,
      title={The Recurrent Transformer: Greater Effective Depth and Efficient Decoding}, 
      author={Costin-Andrei Oncescu and Depen Morwani and Samy Jelassi and Alexandru Meterez and Mujin Kwun and Sham Kakade},
      year={2026},
      eprint={2604.21215},
      archivePrefix={arXiv},
      primaryClass={cs.LG},
      url={https://arxiv.org/abs/2604.21215}, 
      note={arXiv:2604.21215 [cs.LG]},
}

@inproceedings{Ke2018,
 author = {Ke, Nan Rosemary and ALIAS PARTH GOYAL, Anirudh Goyal and Bilaniuk, Olexa and Binas, Jonathan and Mozer, Michael C and Pal, Chris and Bengio, Yoshua},
 booktitle = {Advances in Neural Information Processing Systems},
 editor = {S. Bengio and H. Wallach and H. Larochelle and K. Grauman and N. Cesa-Bianchi and R. Garnett},
 pages = {},
 publisher = {Curran Associates, Inc.},
 title = {Sparse Attentive Backtracking: Temporal Credit Assignment Through Reminding},
 url = {https://proceedings.neurips.cc/paper_files/paper/2018/file/e16e74a63567ecb44ade5c87002bb1d9-Paper.pdf},
 volume = {31},
 year = {2018}
}

@inproceedings{Yang2024gatedlinearattention,
author = {Yang, Songlin and Wang, Bailin and Shen, Yikang and Panda, Rameswar and Kim, Yoon},
title = {Gated linear attention transformers with hardware-efficient training},
year = {2024},
publisher = {JMLR.org},
booktitle = {Proceedings of the 41st International Conference on Machine Learning},
articleno = {2333},
numpages = {23},
location = {Vienna, Austria},
series = {ICML'24}
}

@inproceedings{
siems2025deltaproduct,
title={DeltaProduct: Improving State-Tracking in Linear {RNN}s via Householder Products},
author={Julien Siems and Timur Carstensen and Arber Zela and Frank Hutter and Massimiliano Pontil and Riccardo Grazzi},
booktitle={The Thirty-ninth Annual Conference on Neural Information Processing Systems},
year={2025},
url={https://openreview.net/forum?id=SoRiaijTGr}
}

@misc{baldelli2026,
      title={LLMs Can't Play Hangman: On the Necessity of a Private Working Memory for Language Agents}, 
      author={Davide Baldelli and Ali Parviz and Amal Zouaq and Sarath Chandar},
      year={2026},
      eprint={2601.06973},
      archivePrefix={arXiv},
      primaryClass={cs.CL},
      url={https://arxiv.org/abs/2601.06973}, 
      note={arXiv:2601.06973 [cs.CL]}
}

@misc{chen2026,
      title={Training-Free Looped Transformers}, 
      author={Lizhang Chen and Jonathan Li and Chen Liang and Ni Lao and Qiang Liu},
      year={2026},
      eprint={2605.23872},
      archivePrefix={arXiv},
      primaryClass={cs.LG},
      url={https://arxiv.org/abs/2605.23872}, 
      note={arXiv:2605.23872 [cs.CL]}
}

@misc{sanyal2026,
  author = {Sunny Sanyal},
  title = {Looped-GPT: Looping During Pre-training improves Generalization},
  year = {2026},
  publisher = {Blog},
  url = {https://sanyalsunny111.github.io/posts/2026-01-15-post1-looped-gpt/}
}

@misc{fagnou2026,
      title={Trading Complexity for Expressivity Through Structured Generalized Linear Token Mixing}, 
      author={Erwan Fagnou and Paul Caillon and Blaise Delattre and Alexandre Allauzen},
      year={2026},
      publisher = {JMLR.org},
      booktitle = {Proceedings of the 43d International Conference on Machine Learning},
      numpages = {25},
      location = {Seoul, S. Korea},
      series = {ICML'26}
}
\end{document}